\documentclass[12pt,twoside]{article}
\usepackage{latexsym}
\newdimen\paravsp  \paravsp=1.3ex
\def\,{\mskip 3mu} \def\>{\mskip 4mu plus 2mu minus 4mu} \def\;{\mskip 5mu plus 5mu} \def\!{\mskip-3mu}
\def\dispmuskip{\thinmuskip= 3mu plus 0mu minus 2mu \medmuskip=  4mu plus 2mu minus 2mu \thickmuskip=5mu plus 5mu minus 2mu}
\def\textmuskip{\thinmuskip= 0mu                    \medmuskip=  1mu plus 1mu minus 1mu \thickmuskip=2mu plus 3mu minus 1mu}
\textmuskip
\def\beqn{\dispmuskip\begin{displaymath}}\def\eeqn{\end{displaymath}\textmuskip}

\def\abstract#1{\centerline{\bf\small
Abstract}\begin{quote}\vspace{-1ex}\small #1 \par\end{quote}\vskip 1ex}
\def\keywords#1{\centerline{\bf\small
Keywords}\begin{quote}\vspace{-1ex}\small #1 \par\end{quote}\vskip 1ex}
\def\paradot#1{\vspace{\paravsp plus 0.5\paravsp minus 0.5\paravsp}\noindent{\bf\boldmath{#1.}}}
\def\paranodot#1{\vspace{\paravsp plus 0.5\paravsp minus 0.5\paravsp}\noindent{\bf\boldmath{#1}}}
\def\req#1{(\ref{#1})}
\def\SetR{I\!\!R}
\def\SetN{I\!\!N}
\def\itb{\bf}                           
\def\eps{\varepsilon}                   
\def\nq{\hspace{-1em}}
\def\l{{\ell}}                          
\def\M{{\cal M}}                        
\def\X{{\cal X}}                        
\def\A{{\cal A}}                        
\def\O{{\cal O}}                        

\def\IZP{\ref{secInduct}\ref{IZP}}
\def\IBR{\ref{secInduct}\ref{IBR}}
\def\IGRUE{\ref{secInduct}\ref{IGRUE}}
\def\IRI{\ref{secInduct}\ref{IRI}}
\def\IOEUP{\ref{secInduct}\ref{IOEUP}}
\def\IOP{\ref{secInduct}\ref{IOP}}
\def\PSB{\ref{secPredict}\ref{PSB}}
\def\PNTM{\ref{secPredict}\ref{PNTM}}
\def\PMLC{\ref{secPredict}\ref{PMLC}}
\def\PGMCC{\ref{secPredict}\ref{PGMCC}}
\def\PLCBDM{\ref{secPredict}\ref{PLCBDM}}
\def\PAIXI{\ref{secPredict}\ref{PAIXI}}
\def\OMUO{\ref{secOptim}\ref{OMUO}}
\def\OLEC{\ref{secOptim}\ref{OLEC}}
\def\OGBM{\ref{secOptim}\ref{OGBM}}
\def\OIA{\ref{secOptim}\ref{OIA}}
\def\SAIXI{\ref{secStruct}\ref{SAIXI}}
\def\SPD{\ref{secStruct}\ref{SPD}}
\def\SAEA{\ref{secStruct}\ref{SAEA}}
\def\SARV{\ref{secStruct}\ref{SARV}}
\def\CPCUH{\ref{secConv}\ref{CPCUH}}
\def\CRL{\ref{secConv}\ref{CRL}}
\def\CRNCE{\ref{secConv}\ref{CRNCE}}
\def\CGT{\ref{secConv}\ref{CGT}}
\def\CPOST{\ref{secConv}\ref{CPOST}}
\def\DGSPM{\ref{secDefine}\ref{DGSPM}}
\def\DPPM{\ref{secDefine}\ref{DPPM}}
\def\DEE{\ref{secDefine}\ref{DEE}}

\begin{document}
\title{\vspace{-4ex}
\vskip 2mm\bf\Large\hrule height5pt \vskip 4mm
Open Problems in Universal \\ Induction \& Intelligence
\vskip 4mm \hrule height2pt}
\author{{\bf Marcus Hutter}\\[3mm]
\normalsize RSISE$\,$@$\,$ANU and SML$\,$@$\,$NICTA \\
\normalsize Canberra, ACT, 0200, Australia \\
\normalsize \texttt{marcus@hutter1.net \ \  www.hutter1.net}
}
\date{21 June 2009}
\maketitle
\vspace*{-4ex}

\abstract{
Specialized intelligent systems can be found everywhere: finger
print, handwriting, speech, and face recognition, spam filtering,
chess and other game programs, robots, et al. This decade the first
presumably complete {\em mathematical} theory of artificial
intelligence based on universal induction-prediction-decision-action
has been proposed. This information-theoretic approach solidifies
the foundations of inductive inference and artificial intelligence.
Getting the foundations right usually marks a significant progress
and maturing of a field. The theory provides a gold standard and
guidance for researchers working on intelligent algorithms. The
roots of universal induction have been laid exactly half-a-century
ago and the roots of universal intelligence exactly one decade ago.
So it is timely to take stock of what has been achieved and what
remains to be done. Since there are already good recent surveys, I
describe the state-of-the-art only in passing and refer the reader
to the literature. This article concentrates on the open problems in
universal induction and its extension to universal intelligence.
\def\contentsname{\centering\normalsize Contents}
{\parskip=-2.7ex\tableofcontents}\vspace*{-4ex}
}


\begin{quote}\it
``The mathematician is by now accustomed to intractable equations,
and even to unsolved problems, in many parts of his discipline.
However, it is still a matter of some fascination to realize
that there are parts of mathematics where the very construction
of a precise mathematical statement of a verbal problem is
itself a problem of major difficulty.'' \par
\hfill --- {\sl Richard Bellman, Adaptive Control Processes (1961) p.194}
\end{quote}

\section{Introduction}\label{secIntro}

What is a good model of the weather changes? %
Are there useful models of the world economy? %
What is the true regularity behind the number sequence 1,4,9,16,...? %
What is the correct relationship between mass, force,
and acceleration of a physical object? %
Is there a causal relation between interest rates and inflation? %
Are models of the stock market purely descriptive or do they have any predictive power? %

\paradot{Induction}
The questions above look like a set of unrelated inquires. What
they have in common is that they seem to be amenable to scientific
investigation. They all ask about a model for or relation between
observations. The purpose seems to be to explain or understand the
data. Generalizing from data to general rules is called {\em inductive
inference}, a core problem in philosophy
\cite{Hume:1739,Popper:34,Howson:03} and a key task of science
\cite{Levi:74,Earman:93,Wallace:05}.

But why do or should we care about modeling the world? Because this
is what science is about \cite{Salmon:06}? As indicated above,
models should be good, useful, true, correct, causal, predictive, or
descriptive \cite{Frigg:06}. Digging deeper, we see that models are
mostly used for prediction in related but new situations, especially
for predicting future events \cite{Wikipedia:08pm}.

\paradot{Predictions}
Consider the apparently only slight variation of the questions above: %
What is the correct answer in an IQ test asking to
continue the sequence 1,4,9,16,...? %
Given historic stock-charts, can one predict the quotes of tomorrow? %
Or questions like: Assuming the sun rose every day for 5000 years,
how likely is doomsday (that the sun will not rise) tomorrow? %
What is my risk of dying from cancer next year? %

These questions are instances of the important problem of
time-series forecasting, also called sequence prediction
\cite{Brockwell:02,Cesa:06}.
While inductive inference is about finding models or hypotheses that
{\em explain} the data (whatever {\em explain} actually shall mean),
{\em prediction} is concerned about forecasting the future.
Finding models is interesting and
useful, since they usually help us to (partially) answer such {\em
predictive} questions \cite{Geisser:93,Chatfield:03}.
While the usefulness of predictions is clearer to the layman than
the purpose of the scientific inquiry for models, one may again ask,
why we do or should we care about making predictions?

\paradot{Decisions}
Consider the following questions: Shall I take my umbrella or wear
sunglasses today? Shall I invest my assets in stocks or bonds? Shall
I skip work today because it might be my last day on earth? Shall I
irradiate or remove the tumor of my patient? These questions ask for
decisions that have some (minor to drastic) consequences. We usually
want to make ``good'' decisions, where the quality is measured in
terms of some reward (money, life expectancy) or loss
\cite{Ferguson:67,DeGroot:70,Jeffrey:83}. In order to compute this
reward as a function of our decision, we need to predict the
environment: whether there will be rain or sunshine today, whether
the market will go up or down, whether doomsday is tomorrow, or
which type of cancer the patient has. Often forecasts are uncertain
\cite{Paris:95}, but this is still better than no prediction.
Once we arrived at a (hopefully good) decision, what do we do next?

\paradot{Actions}
The obvious thing is to execute the decision, i.e.\ to perform some
action consistent with the decision arrived at. The action may not
influence the environment, like taking umbrella versus sunglasses
does not influence the future weather (ignoring the butterfly
effect) or small stock trades. These settings are called passive
\cite{Hutter:03optisp}, and the action part is of marginal
importance and usually not discussed. On the other hand, a patient
might die from a wrong treatment, or a chess player loses a figure
and possibly the whole game by making one mistake. These settings
are called (re)active \cite{Hutter:07aixigentle}, and their analysis is
immensely more involved than the passive case \cite{Bertsekas:06}.

\paranodot{And now?}
There are many theories and algorithms and whole research fields and
communities dealing with some aspects of induction, prediction,
decision, or action. Some of them will be detailed below.
Finding solutions for every particular (new) problem is possible and
useful for many specific applications. Trouble is that this approach
is cumbersome and prone to disagreement or contradiction
\cite{Kemp:03}. Some researchers feel that this is the nature of
their discipline and one can do little about it \cite{Kellert:06}.
But in science (in particular math, physics, and computer science)
previously separate approaches are constantly being unified towards
more and more powerful theories and
algorithms \cite{Green:87,Greene:00}.
There is at least one field, where we {\em must} put everything
(induction+prediction+decision+action) together in a completely
formal (preferably elegant) way, namely Artificial Intelligence
\cite{Russell:03}. Such a general and formal theory of AI has been
invented about a decade ago \cite{Hutter:00kcunai}.

\paradot{Contents}
In {\em Section \ref{secUAI}} I will give a brief introduction into this
universal theory of AI. It is based on an unexpected
unification of algorithmic information theory
and sequential decision theory. The corresponding AIXI agent is the
first sound, complete, general, rational agent in any relevant but
unknown environment with reinforcement feedback
\cite{Hutter:04uaibook,Oates:06}. It is likely the best possible
such agent in a sense to be explained below.

{\em Section \ref{secHist}} describes the historic origin of the
AIXI model. One root is Solomonoff's theory \cite{Solomonoff:60} of
{\em universal induction}, which is closely connected to algorithmic
complexity. The other root is Bellman's {\em adaptive control
theory} \cite{Bellman:57} for optimal {\em sequential decision}
making. Both theories are now half-a-century old. From an
algorithmic information theory perspective, AIXI generalizes optimal
passive universal induction to the case of active agents. From a
decision-theoretic perspective, AIXI is a universal Bayes-optimal
learning algorithm.

{\em Sections \ref{secInduct}--\ref{secDefine}} constitute the core
of this article describing the open problems around universal
induction \& intelligence. Most of them are taken from the book
\cite{Hutter:04uaibook} and paper \cite{Hutter:07uspx}. I focus on
questions whose solution has a realistic chance of advancing the
field. I avoid technical open problems whose global significance is
questionable.

Solomonoff's half-a-century-old theory of universal induction is
already well developed. Naturally, most remaining open
problems are either philosophically or technically deep.

Its generalization to Universal Artificial Intelligence seems to
be quite intricate. While the AIXI model itself is very elegant,
its analysis is much more cumbersome. Although AIXI has been
shown to be optimal in some senses, a convincing notion
of optimality is still lacking. Convergence results also exist, but
are much weaker than in the passive case.

Its construction makes it plausible that AIXI is the optimal
rational general learning agent, but unlike the induction case,
victory cannot be claimed yet. It would be natural, hence, to
compare AIXI to alternatives, if there were any. Since there are
no competitors yet, one could try to create some. Finally, AIXI is only
``essentially'' unique, which gives rise to some more open
questions.

Given that AI is about designing intelligent systems, a serious
attempt should be made to {\em formally} define intelligence in the
first place. Astonishingly there have been not too many
attempts. There is one definition that is closely related to AIXI,
but its properties have yet to be explored.

The final {\em Section \ref{secConc}} briefly discusses the flavor,
feasibility, difficulty, and interestingness of the raised
questions, and takes a step back and briefly compares the
information-theoretic approach to AI discussed in this article to
others.

\section{Universal Artificial Intelligence}\label{secUAI}

\paradot{Artificial Intelligence}
The science of artificial intelligence (AI) may be defined as the
construction of intelligent systems ({\em artificial
agents}) and their analysis \cite{Russell:03}. A natural definition
of a {\em system} is anything that has an input and an output
stream, or equivalently an agent that acts and observes.
{\em Intelligence} is more complicated. It can have many faces like
{\em creativity}, {\em solving problems}, {\em pattern recognition},
{\em classification}, {\em learning}, {\em induction}, {\em
deduction}, {\em building analogies}, {\em optimization}, {\em
surviving in an environment}, {\em language processing}, {\em
planning}, and {\em knowledge acquisition and processing}.
Informally, {\em AI is concerned with developing agents that
perform well in a large range of environments} \cite{Hutter:07iorx}.
A formal definition incorporating every aspect of intelligence,
however, seems difficult.
In order to solve this problem we need to solve the induction,
prediction, decision, and action problem, which seems like
a daunting (some even claim impossible) task:
Intelligent {\em actions} are based on informed {\em decisions}.
Attaining good decisions requires {\em predictions} which are
typically based on models of the environments. Models are
constructed or learned from past observations via {\em
induction}.
Fortunately, based on the {\em deep philosophical insights} and
{\em powerful mathematical developments} listed in Section
\ref{secHist}, these problems have been overcome, at least in
theory.

\paradot{Universal Artificial Intelligence (UAI)}
Most, if not all, known facets of intelligence can be formulated as
goal driven or, more precisely, as maximizing some reward or utility
function. It is, therefore, sufficient to study goal-driven AI;
e.g.\ the (biological) goal of animals and humans is to survive and
spread. The goal of AI systems should be to be useful to humans. The
problem is that, except for special cases, we know neither the
utility function nor the environment in which the agent will operate
in advance.
What do we need (from a mathematical point of view) to construct a
universal optimal learning agent interacting with an arbitrary
unknown environment? The theory, coined {\em AIXI}, developed in
this decade and explained in \cite{Hutter:04uaibook} says: {\em All
you need is} {\em Occam} \cite{Franklin:02}, {\em Epicurus}
\cite{Asmis:84}, {\em Turing} \cite{Turing:36}, {\em Bayes}
\cite{Bayes:1763}, {\em Solomonoff} \cite{Solomonoff:64}, {\em
Kolmogorov} \cite{Kolmogorov:65}, and {\em Bellman}
\cite{Bellman:57}:
Sequential decision theory \cite{Bertsekas:06} ({\em Bellman}'s
equation) formally solves the problem of rational agents in
uncertain worlds if the true environmental probability distribution
is known. If the environment is unknown, {\em Bayes}ians
\cite{Berger:93} replace the true distribution by a weighted mixture
of distributions from some (hypothesis) class. Using the large class
of all (semi)measures that are (semi)computable on a {\em Turing}
machine bears in mind {\em Epicurus}, who teaches not to discard
any (consistent) hypothesis. In order not to ignore {\em Occam},
who would select the simplest hypothesis, {\em Solomonoff} defined
a universal prior that assigns high/low prior weight to
simple/complex environments, where {\em Kolmogorov} quantifies
complexity \cite{Hutter:07ait,Li:08}.
All other concepts and phenomena attributed to intelligence are emergent.
All together, this solves all conceptual problems
\cite{Hutter:04uaibook}, and ``only'' computational problems remain.

\paradot{Kolmogorov complexity}
Kolmogorov \cite{Kolmogorov:65} defined the complexity of a string
$x\in\X^*$ over some finite alphabet $\X$ as the length $\ell$ of a
shortest description $p\in\{0,1\}^*$ on a universal Turing machine $U$:
\beqn
  \mbox{Kolmogorov complexity:}\qquad K(x):=\min_p\{\ell(p):U(p)=x\}
\eeqn
A string is simple if it can be described by a short program, like
``the string of one million ones'', and is complex if there is no
such short description, like for a random string whose shortest
description is specifying it bit-by-bit.
For non-string objects $o$ one defines
$K(o):=K(\langle o\rangle)$, where $\langle o\rangle\in\X^*$ is
some standard code for $o$.
Kolmogorov complexity \cite{Kolmogorov:65,Hutter:08kolmo} is a key
concept in (algorithmic) information theory \cite{Li:08}. An
important property of $K$ is that it is nearly independent of the
choice of $U$, i.e.\ different choices of $U$ change $K$ ``only'' by
an additive constant (see Section \PNTM). Furthermore it leads to
shorter codes than any other effective code. $K$ shares many
properties with Shannon's entropy (information measure) $S$
\cite{MacKay:03,Cover:06}, but $K$ is superior to $S$ in many
respects. Foremost, $K$ measures the information of individual
outcomes, while $S$ can only measure expected information of random
variables. To be brief, $K$ is an excellent universal complexity
measure, suitable for quantifying Occam's razor. The major drawback
of $K$ as complexity measure is its incomputability. So in practical
applications it has always to be approximated, e.g.\ by Lempel-Ziv
compression \cite{Lempel:76,Cilibrasi:05}, or by CTW
\cite{Willems:97} compression, or by using two-part codes like in
MDL and MML, or by others.

\paradot{Solomonoff induction}
Solomonoff \cite{Solomonoff:64} defined (earlier) the closely
related universal a priori probability $M(x)$ as the probability
that the output of a universal (monotone) Turing machine $U$ starts
with $x$ when provided with fair coin flips on the input tape
\cite{Hutter:07algprob}. Formally,
\beqn
  \mbox{Solomonoff prior:}\qquad\qquad M(x):=\sum_{p:U(p)=x*}2^{-\ell(p)},
\eeqn
where the sum is over all (possibly non-halting) so-called minimal
programs $p$ which output a string starting with $x$. Since the sum
is dominated by short programs, we have $M(x)\approx 2^{-K(x)}$
(formally $-\log M(x)=K(x)+O(\log\l(x))$), i.e.\ simple/complex
strings are assigned a high/low a-priori probability.
A different representation is as follows \cite{Zvonkin:70}:
Let $\M=\{\nu\}$ be a countable class of probability measures $\nu$
(environments) on infinite sequences $\X^\infty$, $\mu\in\M$ be the
true sampling distribution, i.e.\ $\mu(x)$ is the true probability
that an infinite sequences starts with $x$, and
$\xi_\M(x):=\sum_{\nu\in\M} w_\nu\nu(x)$ be the $w$-weighted average
called Bayesian mixture distribution. One can show that
$M(x)=\xi_{\M_U}(x)$, where $\M_U$ includes all computable
probability measures and $w_\nu=2^{-K(\nu)}$. More precisely,
$\M_U:=\{\nu_1,\nu_2,...\}$ consists of an effective enumeration of
all so-called lower semi-computable semi-measures $\nu_i$, and
$K(\nu_i):=K(i):=K(\langle i\rangle)$ \cite{Li:08}.

$M$ can be used as a universal sequence predictor, which
outperforms in a strong sense all other predictors.
Consider the classical online sequence prediction task: Given
$x_{<t}\equiv x_{1:t-1}:=x_1...x_{t-1}$, predict $x_t$; then observe
the true $x_t$; $t\leadsto t+1$; repeat. For $x_{1:\infty}$ generated
by the unknown ``true'' distribution $\mu\in\M_U$,
one can show \cite{Solomonoff:78} that the universal predictor
$M(x_t|x_{<t}):=M(x_{1:t})/M(x_{<t})$ rapidly converges to the true
probability $\mu(x_t|x_{<t})=\mu(x_{1:t})/\mu(x_{<t})$ of the next
observation $x_t\in\X$ given history $x_{<t}$. That is, $M$ serves
as an excellent predictor of any sequence sampled from any
computable probability distribution.

\paradot{The AIXI model}
\def\nc{}
\def\ac{}
\def\oc{}
\def\rc{}
\def\xc{}
\def\ec{}
\def\uc{}
\def\ic{}
It is possible to write down the {\ic AIXI} model explicitly in one
line \cite{Hutter:07aixigentle}, although one should not expect to
be able to grasp the full meaning and power from this compact
representation.

{\ic AIXI} is an {\ac agent} that interacts with an {\ec environment} in
cycles $k=1,2,...,m$. In cycle $k$, {\ic AIXI} takes {\ac action
$a_k$} (e.g.\ a limb movement) based on past perceptions $\oc o_1\rc
r_1..\oc o_{k-1}\rc r_{k-1}$ as defined below. Thereafter, the {\ec
environment} provides a (regular) {\oc observation $o_k$} (e.g.\ a
camera image) to {\ic AIXI} and a real-valued {\rc reward $r_k$}.
The {\rc reward} can be very scarce, e.g.\ just +1 (-1) for winning
(losing) a chess game, and 0 at all other times. Then the next cycle
$k+1$ starts. Given the above, AIXI is defined by:
\beqn
  \mbox{\ic AIXI:}\qquad \ac a_k \nc\;:=\; \ac\arg\max_{a_k}\xc\sum_{\oc o_k\rc r_k} 
  \nc \ldots \ac\max_{a_m}\xc\sum_{\oc o_m\rc r_m}
  \rc[r_k+\cdots+r_m]
  \uc\nq\nq\nq\!\!\!\sum_{{\ec q}\,:\,U({\ec q},{\ac a_1..a_m})={\oc o_1\rc r_1..\oc o_m\rc r_m}}\nq\nq\nq\!\!\! 2^{-\ell({\ec q})}
\eeqn
The expression shows that {\ic AIXI} tries to {\ac max}imize its
{\rc total future reward $r_k+...+r_m$}. If the environment is
modeled by a deterministic {\ec program $q$}, then the future {\xc
perceptions} $\rc...{\oc o_k\rc r_k..\oc o_m\rc r_m} = \uc U({\ec
q},{\ac a_1..a_m})$ can be computed, where $\uc U$ is a {\uc
universal (monotone Turing) machine} executing $\ec q$ given $\ac
a_1..a_m$. Since $\ec q$ is unknown, {\ic AIXI} has to maximize its
{\xc expected} reward, i.e.\ average $\rc r_k+...+r_m$ over all
possible {\ec perceptions} created by all possible environments $\ec
q$. The simpler an environment, the higher is its a-priori
contribution $\uc 2^{-\ell({\ec q})}$, where simplicity is measured
by the {\uc length $\ell$} of program $\ec q$. The inner sum
$\sum_{q:...}2^{-\ell(q)}$ generalizes Solomonoff's a-priori
distribution $M$ by including actions. Since {\ec noisy
environments} are just mixtures of deterministic environments, they
are automatically included.
The sums in the formula constitute the averaging process. {\xc
Averaging} and {\ac maximization} have to be performed in
chronological order, hence the interleaving of {\ac max} and
$\xc\Sigma$ (similarly to minimax for games).
The {\em value} $V$ of AIXI (or any other agent) is its expected
reward sum.

One can fix any finite action and perception space, any reasonable
$\uc U$, and any large finite lifetime $m$. This completely and
uniquely defines {\ic AIXI}'s actions $\ac a_k$, which
are limit-computable via the expression above (all quantities are
known).

{\em That's it!} Ok, not really. It takes a whole book and more to
explain why {\ic AIXI} likely is the most intelligent
general-purpose agent and incorporates all aspects of rational
intelligence. In practice, {\ic AIXI} needs to be approximated. {\ic
AIXI} can also be regarded as the gold standard which other
practical general purpose AI programs should aim at (analogue to
minimax approximations/heuristics).

\paradot{The role of AIXI for AI}
The AIXI model can be regarded as the first {\em complete theory of
AI}. Most if not all AI problems can easily be formulated within
this theory, which reduces the conceptual problems to pure
computational questions. Solving the conceptual part of a problem
often causes a quantum leap forward in a field. Two analogies may
help: QED is a {\em complete} theory of {\em all} chemical
processes. ZFC solved the {\em conceptual} problems of sets (e.g.\
Russell's paradox).

From an algorithmic information theory (AIT) perspective, the AIXI
model generalizes optimal passive universal induction to the case of
active agents. From a decision-theoretic perspective, AIXI is a
suggestion of a new (implicit) ``learning'' algorithm, which may
overcome all (except computational) problems of previous
reinforcement learning algorithms. If the optimality theorems of
universal induction and decision theory generalize to the unified
AIXI model, we would have, for the first time, a universal
(parameterless) model of an optimal rational agent in any computable
but unknown environment with reinforcement feedback.

Although deeply rooted in algorithm theory, AIT mainly neglects
computation time and so does AIXI. It is important to note that this
does not make the AI problem trivial. Playing chess optimally or
solving NP-complete problems become trivial, but driving a car or
surviving in nature do not. This is because it is a challenge itself
to well-define the latter problems, not to mention presenting an
algorithm. In other words: The AI problem has not yet been well
defined (cf.\ the quote after the abstract). One may view AIXI as a
suggestion of such a mathematical definition.

Although Kolmogorov complexity is incomputable in general,
Solomonoff's theory triggered an entire field of research on
computable approximations.  This led to numerous practical
applications \cite{Li:07appait}. If the AIXI model should lead to a
universal ``active'' decision maker with properties analogous to
those of universal ``passive'' predictors, then we could expect a
similar stimulation of research on resource-bounded, practically
feasible variants.  First attempt have been made to test the power
and limitations of AIXI and downscaled versions like AIXI$tl$ and
AI$\xi$ \cite{Hutter:06aixifoe,Pankov:08}, as well as related models
derived from basic concepts of algorithmic information theory.

So far, some remarkable and surprising results have already been
obtained (see Section \ref{secHist}). A 2, 12, 60, 300 page
introduction to the AIXI model can be found in
\cite{Hutter:01decision,Hutter:01aixi,Hutter:07aixigentle,Hutter:04uaibook},
respectively, and a gentle introduction to UAI in \cite{Legg:08}.

\section{History and State-of-the-Art}\label{secHist}

The theory of UAI and AIXI build on the theories of universal
induction, universal prediction, universal decision making,
and universal agents.
From a historical and research-field perspective, the {\em AIXI
model} is based on two otherwise unconnected fundamental theories:
\begin{itemize} 
\item[(1)] The major basis is {\it Algorithmic information theory}
\cite{Li:08}, initiated by
\cite{Solomonoff:64,Kolmogorov:65,Chaitin:66}, which builds the
foundation of complexity and randomness of individual objects. It
can be used to quantify Occam's razor principle
(use the simplest theory consistent with the data). This in turn
allowed {\em Solomonoff} to come up with a {\em universal theory of
induction} \cite{Solomonoff:64,Solomonoff:78}.

\item[(2)] The other basis is the theory of optimal {\em sequential decisions},
initiated by Von Neumann \cite{VonNeumann:44} and Bellman
\cite{Bellman:57}. This theory builds the basis of modern {\em
reinforcement learning} \cite{Sutton:98}.
\end{itemize}
This section outlines the history and state-of-the-art of the
theories and research fields involved in the AIXI model.

\paradot{Algorithmic information theory (AIT)}
In the 1960's \cite{Kolmogorov:65,Solomonoff:64,Chaitin:66}
introduced a new machine independent complexity measure for
arbitrary computable data. The Kolmogorov complexity $K(x)$ is
defined as the length of the shortest program on a universal Turing
machine that computes $x$. It is closely related to Solomonoff's
universal a-priori probability $M(x)\approx 2^{-K(x)}$ (see above),
Martin-L\"of randomness of individual sequences \cite{MartinLoef:66},
time-bounded complexity \cite{Levin:84}, universal optimal search
\cite{Levin:73search}, the speed prior \cite{Schmidhuber:02speed},
the halting probability $\Omega$ \cite{Chaitin:87}, strong mathematical
undecidability \cite{Chaitin:03}, generalized
probability and complexity \cite{Schmidhuber:02gtm}, algorithmic
statistics \cite{Gacs:01,Vereshchagin:02,Vitanyi:02}, and others.

Despite its incomputability, AIT found many applications
in philosophy, practice, and science:
The minimum message/description length (MML/MDL) principles
\cite{Wallace:68,Rissanen:78,Rissanen:89} can be regarded as a
practical approximation of Kolmogorov complexity. MML\&MDL
are widely used in machine learning applications
\cite{Quinlan:89,Gao:89,Jurka:93,Pednault:89,Wallace:05,Gruenwald:07book}.
The latest, most direct and impressive applications are via the
universal similarity metric \cite{Cilibrasi:05,Cilibrasi:06}.
Schmidhuber produced another range of impressive applications to %
neural networks \cite{Schmidhuber:97nn,Schmidhuber:97bias}, %
in search problems \cite{Schmidhuber:04oops}, %
and even in the fine arts \cite{Schmidhuber:97art}.
By carefully approximating Kolmogorov complexity, AIT sometimes lead
to results unmatched by other approaches.
Besides these practical applications, AIT is used to simplify proofs
via the incompressibility method, improves Shannon information, is
used in reversible computing, physical entropy and Maxwell daemon
issues, artificial intelligence, and the asymptotically fastest algorithm
for all well-defined problems
\cite{Calude:02,Hutter:04uaibook,Hutter:02fast,Hutter:07ait,Li:08}.

\paradot{Universal Solomonoff induction}
How and in which sense induction is possible at all has been subject
to long philosophical controversies
\cite{Hume:1739,Stork:01,Hutter:04uaibook}. Highlights are
Epicurus' principle of multiple explanations \cite{Asmis:84},
Occam's razor (simplicity) principle \cite{Franklin:02}, and Bayes'
rule for conditional probabilities \cite{Bayes:1763,Earman:93}.
Solomonoff \cite{Solomonoff:64} elegantly unified these aspects with
the concept of universal Turing machines \cite{Turing:36} to one
formal theory of inductive inference based on a universal
probability distribution $M$, which is closely related to Kolmogorov
complexity $K$ ($M(x)\approx 2^{-K(x)}$). The theory allows for
optimally predicting sequences without knowing their true generating
distribution $\mu$ \cite{Solomonoff:78}, and presumably solves the
induction problem. The theory remained for more than 20 years at
this stage, till the work on AIXI started, which resulted in a
beautiful elaboration and extension of Solomonoff's theory.

Meanwhile, the (non)existence of universal priors for several
generalized computability concepts
\cite{Schmidhuber:02gtm,Hutter:03unipriors,Hutter:06unipriorx} has
been classified, rapid convergence of $M$ to the unknown true
environmental distribution $\mu$ \cite{Hutter:01alpha} and tight
error \cite{Hutter:01errbnd} and loss bounds for arbitrary bounded
loss functions and finite alphabet
\cite{Hutter:01loss,Hutter:03spupper} have been proven, and (Pareto)
optimality of $M$ \cite{Hutter:03optisp,Hutter:03unipriors} has been
shown, exemplified on games of chance and compared to predictions
with expert advice \cite{Hutter:03optisp,Hutter:04bayespea}. The
bounds have been further improved by introducting a version of
Kolmogorov complexity that is monotone in the condition
\cite{Hutter:05postbnd,Hutter:07postbndx}. Similar but necessarily
weaker non-asymptotic bounds for universal deterministic/one-part
MDL \cite{Hutter:03unimdl,Hutter:06unimdlx} and discrete two-part
MDL
\cite{Hutter:04mdl2p,Hutter:05mdl2px,Hutter:04mdlspeed,Hutter:06mdlspeedx}
have also been proven. Quite unexpectedly \cite{Hutter:03mlconv} $M$
does {\em not} converge on all Martin-L\"of random sequences
\cite{Hutter:04mlconvx}, but there is a sophisticated remedy
\cite{Hutter:07mlconvxx}.

All together this shows that Solomonoff's induction scheme
represents a universal (formal, but incomputable) solution to all
{\em passive} prediction problems. The most recent studies
\cite{Hutter:06usp} suggest that this theory could solve the
induction problem at whole, or at least constitute a significant
progress in this fundamental problem \cite{Hutter:07uspx}.

\paradot{Sequential decision theory}
Sequential decision theory provides a framework for finding optimal
reward-maximizing strategies in {\em reactive} environments (e.g.\
chess playing as opposed to weather forecasting), assuming the
environmental probability distribution $\mu$ is known.
The Bellman equations \cite{Bellman:57} are at the heart of
sequential decision theory
\cite{VonNeumann:44,Michie:66,Russell:03}. The book
\cite{Bertsekas:06} summarizes open problems and progress in
infinite horizon problems.
%
Sequential decision theory can deal with actions and observations
depending on arbitrary past events. This general setup has been
called AI$\mu$ model in \cite{Hutter:04uaibook,Hutter:07aixigentle}.
Optimality of AI$\mu$ is obvious by construction. This model reduces
in special cases to a range of known models.

\paradot{Reinforcement learning}
If the true environmental probability distribution $\mu$ or
the reward function are unknown, they need to be learned \cite{Sutton:98}.
This dramatically complicates the problem due to the
exploration$\leftrightarrow$exploitation dilemma
\cite{Berry:85,Duff:02,Hutter:04uaibook,Szita:08}.
In order to attack this intrinsically difficult problem, control
theorists typically confine themselves to linear systems with
quadratic loss function, relevant in the control of (simple)
machines, but irrelevant for AI. There are notable exceptions to this confinement, e.g.\
the book \cite{Kumar:86} on stochastic adaptive control and
\cite{Agrawal:89iid,Agrawal:89mdp}, and an increasing number of
more recent work.
Reinforcement learning (RL) (sometimes associated with temporal
difference learning or neural nets) is the instantiation of
stochastic adaptive control theory \cite{Kumar:86} in the machine
learning community. Current research on RL is vast; the most
important conferences are ICML, COLT, ECML, ALT, and NIPS; the most
important journals are JMLR and MLJ. Some highlights and surveys are
\cite{Samuel:59,Barto:83,Sutton:88,Watkins:89,Watkins:92,Moore:93,Tesauro:94,
Wiering:98,Kearns:99rl,Wiering:99,Baum:99,Koller:00,
Singh:03,Guestrin:03,Hutter:08actoptx,Strehl:07,Szita:08,Ross:08pomdp,Hutter:09phimdp,Hutter:09phidbn}
and
\cite{Kaelbling:96,Kaelbling:98,Sutton:98,Boutilier:99,Bertsekas:06}
respectively.
RL has been applied to a variety of real-world problems,
occasionally with stunning success: Backgammon and Checkers
\cite[Chp.11]{Sutton:98}, helicopter control \cite{Ng:04}, and
others.
Nevertheless, existing learning algorithms are very limited
(typically to Markov domains), and non-optimal --- from the very
outset they are approximate or asymptotic only. Indeed, AIXI is
currently the only general and rigorous mathematical formulation of
the addressed problems.

\paradot{The universal algorithmic agent AIXI}
Reinforcement learning algorithms
\cite{Kaelbling:96,Bertsekas:96,Sutton:98} are usually used in
the case of unknown $\mu$. They can succeed if the state space is
either small or has effectively been made small by generalization
techniques. The algorithms work only in restricted, (e.g.\
Markov) domains, have problems with optimally trading off
exploration versus exploitation, have non-optimal learning rate,
are prone to diverge, or are otherwise ad hoc.

The formal solution proposed in
\cite{Hutter:01aixi,Hutter:04uaibook} is to generalize the
universal probability $M$ to include actions as conditions and
replace $\mu$ by $M$ in the AI$\mu$ model, resulting in the AIXI
model, which is presumably universally optimal. It is quite
non-trivial what can be expected from a universally optimal agent
and to properly interpret or define {\em universal}, {\em optimal},
etc \cite{Hutter:07aixigentle}.
It is known that $M$ converges to $\mu$ also in case of multi-step
lookahead as occurs in the AIXI model \cite{Hutter:04selfoptx}, and
that a variant of AIXI is asymptotically self-optimizing and Pareto
optimal \cite{Hutter:02selfopt,Hutter:04mdp}.

The book \cite{Hutter:04uaibook} gives a comprehensive introduction
and discussion of previous achievements on or related to AIXI,
including a critical review, more open problems, comparison to other
approaches to AI, and philosophical issues.

\paradot{Important environmental classes}
In practice, one is often interested in specific classes of problems
rather than the fully universal setting; for example we might be
interested in evaluating the performance of an algorithm designed
solely for function maximization. A taxonomy of abstract
environmental classes from the mathematical perspective of
interacting chronological systems \cite{Hutter:04env,Legg:08} has
been established. The relationships between Bandit problems, MDP
problems, ergodic MDPs, higher order MDPs, sequence prediction
problems, function optimization problems, strategic games,
classification, and many others are formally defined and explored
therein. The work also suggests new abstract environmental classes
that could be useful from an analytic perspective. In
\cite{Hutter:04uaibook}, each problem class is formulated in its
natural way for known $\mu$, and then a formulation within the
AI$\mu$ model is constructed and their equivalence is shown. Then,
the consequences of replacing $\mu$ by $M$ are considered, and in
which sense the problems are formally solved by AIXI.

\paradot{Computational aspects}
The major drawback of AIXI is that it is incomputable, or more
precisely, only asymptotically computable, which makes a
direct implementation impossible. To overcome this problem, the AIXI model can
be scaled down to a model coined AIXI$tl$, which is still superior
to any other time $t$ and length $l$ bounded agent
\cite{Hutter:01aixi,Hutter:04uaibook}. The computation time of
AIXI$tl$ is of the order $t\cdot 2^l$.
A way of overcoming the large multiplicative constant $2^l$ is
possible at the expense of an (unfortunately even larger) additive
constant. The constructed algorithm builds upon Levin search
\cite{Levin:73search,Gaglio:07}. The algorithm is capable of solving all
well-defined problems $p$ as quickly as the fastest algorithm
computing a solution to $p$, save for a factor of $1+\eps$ and
lower-order additive terms \cite{Hutter:02fast}. The solution
requires an implementation of first-order logic, the definition of a
universal Turing machine within it and a proof theory system.
The algorithm as it is, is only of theoretical
interest, but there are more practical variations
\cite{Schmidhuber:04oops,Schmidhuber:04goedel}.
A different, more limited but more practical scaled-down version (coined AI$\xi$) has been
implemented and applied successfully to 2$\times$2 matrix games like
the notoriously difficult repeated prisoner problem and generalized
variants thereof \cite{Hutter:06aixifoe}.

\def\theenumi{\alph{enumi}}
\def\labelenumi{\theenumi)}
\section{Open Problems in Universal Induction}\label{secInduct}\label{secPredict}

The induction problem is a fundamental problem in philosophy
\cite{Hume:1739,Earman:93} and science \cite{Jaynes:03}.
Solomonoff's model is a promising universal solution of the
induction problem. In \cite{Hutter:07uspx}, an attempt has been made
to collect the most important fundamental philosophical and
statistical problems, regarded as open, and to present arguments and
proofs that Solomonoff's theory overcomes them. Despite the force of
the arguments, they are likely not yet sufficient to convince the
(scientific) world that the induction problem is solved. The
discussion needs to be rolled out much further, say, at least one
generally accessible article per one allegedly open problem. Indeed,
this endeavor might even discover some catch in Solomonoff's theory.
Some problems identified and outlined in \cite{Hutter:07uspx} worth
to investigate in more detail are:
\begin{enumerate}
\item {\itb The zero prior problem}.\label{IZP}
The problem is how to {\em confirm universal hypotheses} like
$H:=$``all balls in some urn (or all ravens) are black''. A natural
model is to assume that balls (or ravens) are drawn randomly from an
infinite population with fraction $\theta$ of black balls (or
ravens) and to assume {\em some} prior {\em density} over
$\theta\in[0;1]$ (a uniform density gives the Bayes-Laplace model).
Now we draw $n$ objects and observe that they are all black. The
problem is that the posterior proability
$P[H|\mbox{black}_1...\mbox{black}_n]\equiv 0$, since the prior
probability $P[H]=P[\theta=1]\equiv 0$. Maher's \cite{Maher:04}
approach does not solve the problem \cite{Hutter:07uspx}.

\item {\itb The black raven paradox}\label{IBR}
by  Carl Gustav Hempel goes as follows \cite[Ch.11.4]{Rescher:01}:
Observing $B$lack $R$avens confirms the hypothesis $H$ that all
ravens are black. In general, $(i)$ hypothesis $R\to B$ is confirmed
by $R$-instances with property $B$. Formally substituting
$R\leadsto\neg B$ and $B\leadsto\neg R$ leads to $(ii)$ hypothesis $\neg
B\to\neg R$ is confirmed by $\neg B$-instances with property $\neg
R$. But $(iii)$ since $R\to B$ and $\neg B\to\neg R$ are logically
equivalent, $R\to B$ must also be confirmed by $\neg B$-instance with
property $\neg R$. Hence by $(i)$, observing Black Ravens confirms
Hypothesis $H$, so by $(iii)$, observing White Socks also confirms
that all Ravens are Black, since White Socks are non-Ravens which
are non-Black. But this conclusion is absurd. Again, neither Maher's
nor any other approach solves this problem.

\item {\itb The Grue problem} \cite{Goodman:83}.\label{IGRUE}
Consider the following two hypotheses: $H1:=$``All emeralds are
green'', and $H2:=$``All emeralds found till year 2020 are green,
thereafter all emeralds are blue''. Both hypotheses are equally well
supported by empirical evidence. Occam's razor seems to favor the
more plausible hypothesis $H1$, but by using new predicates {\em
grue}:=``green till y2020 and blue thereafter'' and {\em
bleen}:=``blue till y2020 and green thereafter'', $H2$ gets simpler
than $H1$.

\item {\itb Reparametrization invariance} \cite{Kass:96}.\label{IRI}
The question is how to extend the symmetry principle from finite
hypothesis classes (all hypotheses are equally likely) to infinite
hypothesis classes. For ``compact'' classes, Jeffrey's
prior \cite{Jeffreys:46} is a solution,  but for non-compact spaces
like $\SetN$ or $\SetR$, classical statistical principles lead to
improper distributions, which are often not acceptable.

\item {\itb Old-evidence/updating problem} and {\em ad-hoc hypotheses} \cite{Glymour:80}.\label{IOEUP}
How shall a Bayesian treat the case when some evidence $E\widehat=x$
(e.g.\ Mercury's perihelion advance) is known well-before the
correct hypothesis/theory/model $H\widehat=\mu$ (Einstein's general
relativity theory) is found? How shall $H$ be added to the Bayesian
machinery a posteriori? What is the prior of $H$? Should it be the
belief in $H$ in a hypothetical counterfactual world in which $E$ is
not known? Can old evidence $E$ confirm $H$? After all, $H$ could
simply be constructed/biased/fitted towards ``explaining'' $E$.
Strictly speaking, a Bayesian needs to choose the hypothesis/model
class before seeing the data, which seldom reflects scientific
practice \cite{Earman:93}.

\item {\itb Other issues/problems}.\label{IOP}
Comparison to Carnap's confirmation theory \cite{Carnap:52} and
Laplace rule \cite{Laplace:1812}, allowing for continuous model
classes, how to incorporate prior knowledge
\cite{Press:02,Goldstein:06}, and others.
\end{enumerate}

Solomonoff's theory has already been intensively studied in the
predictive setting
\cite{Solomonoff:78,Hutter:01errbnd,Hutter:03optisp,Hutter:03spupper,Hutter:07postbndx}
mostly confirming its power, with the occasional unexpected exception
\cite{Hutter:07mlconvxx}. Important open questions are:

\begin{enumerate}\setcounter{enumi}{6}
\item {\itb Prediction of selected bits.}\label{PSB}
Consider a very simple and special case of problem \CRNCE, a
binary sequence that coincides at even times with the preceding
(odd) bit, but is otherwise incomputable. Every child will quickly
realize that the even bits coincide with the preceding odd bit, and
after a while perfectly predict the even bits, given the past bits.
The incomputability of the sequence is no hindrance. It is unknown
whether Solomonoff works or fails in this situation. I expect that
a solution of this special case will lead to general useful insights
and advance this theory (cf.\ problem \CRNCE).

\item {\itb Identification of ``natural'' Turing machines.}\label{PNTM}
In order to pin down the additive/multiplicative constants
that plague most results in AIT, it would be highly desirable
to identify a class of
``natural'' UTMs/USMs which have a variety of favorable
properties. A more moderate approach may be to consider classes
${\cal C}_i$ of universal Turing machines (UTM) or
universal semimeasures (USM) satisfying certain properties ${\cal
P}_i$ and showing that the intersection $\cap_i {\cal C}_i$ is not
empty. Indeed, very occasionally results in AIT only hold
for particular (subclasses of) UTMs \cite{Muchnik:02}.
A grander vision is to find the single ``best'' UTM or USM
\cite{Mueller:06} (a remarkable approach).

\item {\itb Martin-L\"of convergence.}\label{PMLC}
Quite unexpectedly, a loophole in the proof of Martin-L{\"o}f (M.L.)
convergence of $M$ to $\mu$ in the literature has been found
\cite{Hutter:03mlconv}. In \cite{Hutter:04mlconvx} it has been shown
that this loophole cannot be fixed, since M.L.-convergence actually
can fail. The construction of non-universal (semi)measures $D$ and
$W$ that M.L.\ converge to $\mu$ \cite{Hutter:07mlconvxx} partially
rescued the situation. The major problem left open is the
convergence rate for $W\stackrel{M.L.}\longrightarrow\mu$. The
current bound for $D\stackrel{M.L.}\longrightarrow\mu$ is double
exponentially worse than for $M\stackrel{w.p.1}\longrightarrow\mu$.
It is also unknown whether convergence in ratio holds.
Finally, there could still exist {\em universal} semimeasures $M$
(dominating all enumerable semimeasures) for which M.L.-convergence
holds. In case they exist, they probably have particularly
interesting additional structure and properties.

\item {\itb Generalized mixtures and convergence concepts.}\label{PGMCC}
Another interesting and potentially fruitful approach to the above
convergence problem is to consider other classes of semimeasures
$\M$
\cite{Schmidhuber:02speed,Schmidhuber:02gtm,Hutter:03unipriors},
define mixtures $\xi$ over $\M$, and (possibly) generalized
randomness concepts by using this $\xi$ to define a generalized
notion of randomness. Using this approach, in
\cite{Hutter:06unipriorx} it has been shown that convergence holds
for a subclass of Bernoulli distributions if the class is dense, but
fails if the class is gappy, showing that a denseness
characterization of $\M$ could be promising in general. See also
\cite{Hutter:07pquest,Hutter:08pquestx}.

\item {\itb Lower convergence bounds and defect of $M$.}\label{PLCBDM}
One can show that $M(\bar x_t|x_{<t})\geq 2^{-K(t)}$, i.e.\ the
probability of making a wrong prediction $\bar x_t$ converges to
zero slower than any computable summable function. This
shows that, although $M$ converges rapidly to $\mu$ in a
cumulative sense, occasionally, namely for simply describable $t$,
the prediction quality is poor. An easy way to show the lower
bound is to exploit the semimeasure defect of $M$. Do similar
lower bounds hold for a proper (Solomonoff) normalized measure
$M_{norm}$? I conjecture the answer is yes, i.e.\ the lower
bound is not a semimeasure artifact, but ``real''.

\item {\itb Using AIXI for prediction.}\label{PAIXI}
Since AIXI is a unification of sequential decision theory with the
idea of universal probability one may think that the AIXI model for
a sequence prediction problem exactly reduces to Solomonoff's
universal sequence prediction scheme. Unfortunately this is not the
case. For one reason, $M$ is only a probability distribution on the
inputs but not on the outputs. This is also one of the origins of
the difficulty of proving general value bounds for AIXI. The
questions is whether, nevertheless, AIXI predicts sequences as well
as Solomonoff's scheme. A first weak bound in a very restricted
setting has been proven in \cite[Sec.6.2]{Hutter:04uaibook},
showing that progress in this question is possible.
\end{enumerate}
The most important open, but unfortunately likely also the hardest,
problem is the formal identification of natural universal (Turing)
machines \req{PNTM}. A proper solution would eliminate one of the two most
important critiques of the whole field of AIT. Item \req{PAIXI} is an
important question for universal AI.

\section{Open Problems regarding Optimality of AIXI}\label{secOptim}\label{secConv}

AIXI has been shown to be Pareto-optimal and a variant of AIXI to be
self-optimizing \cite{Hutter:02selfopt}. These are important results
supporting the claim that AIXI is universally optimal. More results
can be found in \cite{Hutter:04uaibook}. Unlike the induction case,
the results are not strong enough to alley all doubts.
Indeed, the major problem is not to prove optimality but to come up
with a sufficiently strong but still satisfiable optimality notion
in the reinforcement learning case. The following items list four
potential approaches towards a solution:
\begin{enumerate}
\item {\itb What is meant by universal optimality?}\label{OMUO}
A ``learner'' (like AIXI) may converge to the optimal informed
decision maker (like AI$\mu$) in several senses. Possibly relevant
concepts from statistics are, {\em consistency}, {\em
self-tuningness}, {\em self-optimizingness}, {\em efficiency},
{\em unbiasedness}, {\em asymptotically} or {\em finite
convergence} \cite{Kumar:86}, {\em Pareto-optimality}, and some more
defined in \cite{Hutter:04uaibook}. Some concepts are stronger than
necessary, others are weaker than desirable but suitable to start
with. It is necessary to investigate in more breadth which
properties the AIXI model satisfies.

\item {\itb Limited environmental classes.}\label{OLEC}
The problem of defining and proving general value bounds becomes
more feasible by considering, in a first step, restricted concept
classes. One could analyze AIXI for known classes (like Markov or
factorizable environments) and especially for the new classes ({\it
forgetful}, {\it relevant}, {\it asymptotically learnable}, {\it
farsighted}, {\it uniform}, and {\it (pseudo-)passive)} defined in
\cite{Hutter:04uaibook}.

\item {\itb Generaliztion of AIXI to general Bayes mixtures.}\label{OGBM}
Alternatively one can generalize AIXI to AI$\xi$, where
$\xi(\cdot)=\sum_{\nu\in\M}w_\nu\nu(\cdot)$ is a general
Bayes-mixture of distributions $\nu$ in some class $\M$ and prior
$w_\nu$. If $\M$ is the multi-set of all enumerable semi-measures,
then AI$\xi$ coincides with AIXI. If $\M$ is the (multi)set of
passive semi-computable environments, then AIXI reduces to
Solomonoff's optimal predictor \cite{Hutter:03optisp}. The key is
not to prove absolute results for specific problem classes, but to
prove {\em relative results} of the form ``if there exists a policy
with certain desirable properties, then AI$\xi$ also possesses these
desirable properties''. If there are tasks which cannot be solved by
any policy, AI$\xi$ should not be blamed for failing.

\item {\itb Intelligence Aspects of AIXI.}\label{OIA}
Intelligence can have many faces. As argued in
\cite{Hutter:04uaibook}, it is plausible that AIXI possesses all or
at least most properties an intelligent rational agent should
posses. Some of the following properties could and should be
investigated mathematically: {\it creativity}, {\it problem
solving}, {\it pattern recognition}, {\it classification}, {\it
learning}, {\it induction}, {\it deduction}, {\it building
analogies}, {\it optimization}, {\it surviving in an environment},
{\it language processing}, {\it planning}.
\end{enumerate}
%
Sources of inspiration can be previously proven loss bounds for
Solomonoff sequence prediction generalized to unbounded horizon,
optimality results from the adaptive control literature, and the
asymptotic self-optimizingness results for the related AI$\xi$
model.
Value bounds for AIXI are expected to be, in a sense, weaker
than the loss bounds for Solomonoff induction because the problem
class covered by AIXI is much larger than the class of sequence
prediction problems.

In the same sense as Gittins' solution to the bandit problem and
Laplace' rule for Bernoulli sequences, AIXI may simply be regarded
as (Bayes-)optimal by construction. Even when accepting this ``easy
way out'', the above questions remain significant: Theorems relating
AIXI to AI$\mu$ would no longer be regarded as optimality proofs of
AIXI, but just as how much harder it becomes to operate when $\mu$
is unknown, i.e.\ progress on the items above is simply
reinterpreted.

A weaker goal than to prove optimality of AIXI is to ask for
reasonable convergence properties:

\begin{enumerate}\setcounter{enumi}{5}
\item {\itb Posterior convergence for unbounded horizon.}\label{CPCUH}
Convergence of $M$ to $\mu$ holds somewhat surprisingly even for
unbounded horizon, which is good news for AIXI. 
Unfortunately convergence can be slow, but I expect
that convergence is ``reasonably'' fast for ``slowly'' growing
horizon, which is important in AIXI. It would be useful to quantify
and prove such a result.

\item {\itb Reinforcement learning.}\label{CRL}
Although there is no explicit learning algorithm built into the AIXI
model, AIXI is a reinforcement learning system
capable of receiving and exploiting rewards. The system learns by
eliminating Turing machines $q$ in the definition of $M$ once they
become inconsistent with the progressing history. This is similar to
Gold-style learning \cite{Gold:67}. For Markov
environments (but not for partially observable environments) there
are efficient general reinforcement learning algorithms, like
$TD(\lambda)$ and $Q$ learning. One could
compare the performance (learning speed and quality) of AI$\xi$ to
e.g.\ $TD(\lambda)$ and $Q$ learning, extending
\cite{Hutter:06aixifoe}.

\item {\itb Posterization.}\label{CPOST}
Many properties of Kolmogorov complexity, Solomonoff's prior, and
reinforcement learning algorithms remain valid after
``posterization''. With posterization I mean replacing the total value
$V_{1m}$, the weights $w_\nu$, the complexity $K(\nu)$, the
environment $\nu(or_{1:m}|a_{1:m})$, etc.\ by their ``posteriors''
$V_{km}$, $w_\nu(aor_{<k})$, $K(\nu|aor_{<k})$,
$\nu(or_{k:m}|or_{<k}a_{1:m})$, etc, where $k$ is the current
cycle and $m$ the lifespan of AIXI. Strangely enough for $w_\nu$ chosen as
$2^{-K(\nu)}$ it is not true that $w_\nu(aor_{<k})\sim
2^{-K(\nu|aor_{<k})}$. If this property were true, weak bounds as
the one proven in \cite[Sec.6.2]{Hutter:04uaibook} (which is too
weak to be of practical importance) could be boosted to practical
bounds of order 1. Hence, it is highly import to rescue the
posterization property in some way. It may be valid when grouping
together essentially equal distributions $\nu$.

\item {\itb Relevant and non-computable environments $\mu$.}\label{CRNCE}\label{XXexNComp}
Assume that the observations of AIXI contain irrelevant information,
like noise. Irrelevance can formally be defined as being
statistically independent of future observations and rewards, i.e.\
neither affecting rewards, nor containing information about future
observations. It is easy to see that Solomonoff prediction does not
decline under such noise if it is sampled from a computable
distribution. This likely transfers to AIXI. More interesting is the
case, where the irrelevant input is complex. If it is easily
separable from the useful input it should not affect AIXI. One the
other hand, even in prediction this problem is non-trivial, see
problem \PSB. How robustly does AIXI deal with complex but
irrelevant inputs? A model that explicitly deals with this situation
has been developed in \cite{Hutter:09phimdp,Hutter:09phidbn}.

\item {\itb Grain of truth problem}\label{CGT} \cite{Kalai:93}.
Assume AIXI is used in a multi-agent setup \cite{Weiss:00}
interacting with other agents. For simplicity I only discuss the
case of a single other agent in a competitive setup, i.e.\ a
two-person zero-sum game situation.
We can entangle agents $A$ and $B$ by letting $A$ observe $B$'s
actions and vice versa. The rewards are provided externally by the
rules of the game. The situation where $A$ is AIXI and $B$ is a
perfect minimax player was analyzed in
\cite[Sec.6.3]{Hutter:04uaibook}. In multi-agent systems one is
mostly interested in a symmetric setup, i.e.\ $B$ is also an AIXI.
Whereas both AIXI$s$ {\em may} be able to learn the game and
improve their strategies (to optimal minimax or more generally {\em
Nash equilibrium}), this setup violates one of the basic
assumptions. Since AIXI is incomputable, AIXI($B$) does not
constitute a computable environment for AIXI($A$). More generally,
starting with any class of environments $\M$, then
$\mu\widehat=$AI$\xi_\M$ seems not to belong to class $\M$ for most
(all?) choices of $\M$. Various results can no longer be applied,
since $\mu\not\in\M$ when coupling two AI$\xi$s.
Many questions arise: Are there interesting environmental classes
for which AI$\xi_\M\in\M$ or AI$\xi tl_\M\in\M$? Do AIXI($A/B$)
converge to optimal minimax players? Do AIXI$s$ perform well in
general multi-agent setups?
\end{enumerate}
From the optimality questions above, the first one \req{OMUO} is the
most important, least defined, and likely hardest one: In which
sense can a rational agent in general and AIXI in particular be
optimal?
The multi-agent setting adds another layer of difficulty: The grain
of truth problem \req{CGT} is in my opinion the most important
fundamental problem in game theory and multi-agent systems. Its
satisfactory solution should be worth a Nobel prize or Turing award.

\section{Open Problems regarding Uniqueness of AIXI}\label{secUnique}\label{secStruct}

As a unification of two optimal theories, it is plausible that AIXI
is optimal in the ``union'' of their domains, which has been affirmed
but not finally settled by the positive results derived so far. In
the absence of a definite answer, one should be open to alternative
models, but no convincing competitor exists to date. Most of the
following items describe ideas which, if worked out, might result in
alternative models:

\begin{enumerate}
\item {\itb Action with expert advice.}\label{SAEA}
Expected performance bounds for predictions based on Solomonoff's
prior exist. Inspired by Solomonoff induction, a dual, currently
very popular approach, is ``prediction with expert advice'' (PEA)
\cite{Littlestone:89,Vovk:92,Cesa:06}. Whereas PEA performs well in
any environment, but only relative to a given set of experts,
Solomonoff's predictor competes with {\em any} other predictor, but
only in expectation for environments with computable distribution.
It seems philosophically less compromising to make assumptions on
prediction strategies than on the environment, however weak. PEA has
been generalized to active learning \cite{Hutter:05actexp2,Cesa:06},
but the full reinforcement learning case is still open
\cite{Hutter:06aixifoe}. If successful, it could result in a model
dual to AIXI, but I expect the answer to be negative, which on the
positive side would show the distinguishedness of AIXI. Other ad-hoc
approaches like \cite{Hutter:06actopt,Hutter:08actoptx} are also
unlikely to be competitive.

\item {\itb Actions as random variables.}\label{SARV}
There may be more than one way
for the choice of the generalized $M$
in the AIXI model. For instance, instead of defining
$M$ as in \cite{Hutter:04uaibook} one could treat the agent's
actions $a$ also as universally distributed random variables and
then conditionalize $M$ on $a$.

\item {\itb Structure of AIXI.}\label{SAIXI}
The algebraic properties and the structure of AIXI has barely been
investigated. It is known that the value of AI$\mu$ is a linear
function in $\mu$ and the value of AIXI is a convex function in $\mu$,
but this is neither very deep nor very specific to AIXI.
It should be possible to extract all essentials from AIXI which finally
should lead to an axiomatic characterization of AIXI. The benefit
is as in any {\em axiomatic approach}: It would clearly exhibit the
assumptions, separate the essentials from technicalities,
simplify understanding and, most importantly, guide in finding
proofs.

\item {\itb Parameter dependence.}\label{SPD}
The AIXI model depends on a few parameters: the choice of
observation and action spaces $\O$ and $\A$, the horizon $m$, and
the universal machine $U$. So strictly speaking, AIXI is only
(essentially) unique, if it is (essentially) independent of the
parameters. I expect this to be true, but it has not been proven
yet. The $U$-dependence has been discussed in problem \PNTM.
Countably infinite $\O$ and $\A$ would provide a rich enough
interface for all problems, but even binary $\O$ and $\A$ are
sufficient by sequentializing complex observations and actions. For
special classes one could choose $m\to\infty$ \cite{Bertsekas:06};
unfortunately, the universal environment $M$ does not belong to any
of these special classes. See
\cite{Hutter:04uaibook,Hutter:06discount,Hutter:07iorx} for some
preliminary considerations.
\end{enumerate}

\section{Open Problems in Defining Intelligence}\label{secDefine}

A fundamental and long standing difficultly in the field of
artificial intelligence is that (generic) intelligence itself is not
well defined.
%
It is an anomaly that nowadays most AI researchers avoid discussing
intelligence, which is caused by several factors: It is a difficult
old subject, it is politically charged, it is not necessary for
narrow AI which focusses on specific applications, AI research is
done mainly by computer scientists who mainly care about algorithms
rather than philosophical foundations, and the popular belief that
general intelligence is principally unamenable to a mathematical
definition. These reasons explain but only partially justify the
low effort in trying to define intelligence.

Assume we had a definition, ideally a formal, objective,
non-anthropocentric, and direct method of measuring intelligence, or
at least a very general intelligence-like performance measure that
could serve as an adequate substitute.  This would bring the higher
goals of the field into tight focus and allow us to objectively
compare different approaches and judge the overall progress. Indeed,
formalizing and rigorously defining a previously vague concept
usually constitutes a quantum leap forward in the field: Cf.\
set theory, logical reasoning, infinitesimal calculus, energy,
temperature, etc.
Of course there is (some) work on defining \cite{Hutter:07idefs} and
testing \cite{Hutter:07intest} intelligence (see \cite{Hutter:07iorx}
for a comprehensive list of references):

The famous Turing test \cite{Turing:50,Saygin:00,Loebner:90}
involves human interaction, so is unfortunately informal and
anthropocentric, others are large ``messy'' collections of existing
intelligence tests \cite{Bringsjord:03,Alvarado:00} (``shotgun''
approaches), which are subjective and lack a clear theoretical
grounding, and are potentially too narrow.

There are some more elegant solutions based on classical
\cite{Horst:02} and algorithmic \cite{Chaitin:82} information
theory (``C-Test'' \cite{Hernandez:98, Hernandez:00btt,
Hernandez:00cmi}), the latter closely related to Solomonoff's
\cite{Solomonoff:64} ``perfect'' inductive inference model. The
simple program in \cite{Sanghi:03} reached good IQ scores on some
of the more mathematical tests.

One limitation of the C-Test however is that it only deals with
compression and (passive) sequence prediction, while humans or
machines face reactive environments where they are able to change
the state of the environment through their actions. AIXI generalizes
Solomonoff to reactive environments, which suggested an extremely
{\em general}, {\em objective}, {\em fundamental}, and {\em
formal} performance measure \cite{Hutter:06ior,Legg:08}. This
so-called Intelligence Order Relation (IOR) \cite{Hutter:04uaibook}
even attracted the popular scientific press
\cite{Graham-Rowe:05,Fievet:05}, but the theory surrounding it has
not yet been adequately explored.
Here I only describe three non-technical open problems in defining
intelligence.

\begin{enumerate}
\item {\itb General and specific performance measures.}\label{DGSPM}
Currently it is only partially understood how the IOR theoretically
compares to the myriad of other tests of intelligence such as
conventional IQ tests or even other performance tests proposed by AI
other researchers. Another open question is whether the IOR might in
some sense be too general. One may narrow the IOR to specific
classes of problems \cite{Hutter:04env} and compare how the
resulting IOR measures compare to standard performance
measures for each problem class.  This could shed light on aspects
of the IOR and possibly also establish connections between seemingly
unrelated performance metrics for different classes of problems.

\item {\itb Practical performance measures.}\label{DPPM}
A more practically orientated line of investigation would be to
produce a resource bounded version of the IOR like the one in
\cite[Sec.7]{Hutter:04uaibook}, or perhaps some of its special
cases.  This would allow one to define a practically implementable
performance test, similar to the way in which the C-Test has been
derived from incomputable definitions of compression using $Kt$
complexity \cite{Hernandez:00btt}.  As there are many subtle kinds
of resource bounded complexity \cite{Li:08}, the advantages and
disadvantages of each in this context would need to be carefully
examined.  Another possibility is the recent Speed Prior
\cite{Schmidhuber:02speed} or variants of this approach.

\item {\itb Experimental evaluation.}\label{DEE}
Once a computable version of the IOR had been defined, one
could write a computer program that implements it.  One could
then experimentally explore its characteristics in a range of
different problem spaces.  For example, it might be possible to find
correlations with IQ test scores when applied to humans, like has been
done with the C-Test \cite{Hernandez:98}.  Another possibility would
be to consider more limited domains like classification problems or
sequence prediction problems and to see whether the relative
performance of algorithms according to the IOR agrees with
standard performance measures and real world performance.
\end{enumerate}
A comprehensive collection, discussion and comparison of
verbal and formal intelligence tests, definitions, and measures
can be found in \cite{Hutter:07iorx}.

\section{Conclusions}\label{secConc}

\paradot{The flavor of the open questions}
While most of the key questions about universal sequence prediction
have been solved, many key questions about universal AI remain open
to date.
The questions in Sections \ref{secInduct}-\ref{secDefine} are
centered around the AIT approach to induction and AI, but many
require interdisciplinary working.
A more detailed account with technical details can be found in the
book \cite{Hutter:04uaibook} and paper \cite{Hutter:07uspx}.
Most questions are amenable to a {\em rigorous mathematical}
treatment, including the more philosophically or vaguely sounding
ones. Progress on the latter can achieved in the usual way by
cycling through
$(i)$ craft or improve mathematical definitions that resemble the
intuitive concepts to be studied (e.g.\ ``natural'', ``generalization'',
``optimal''), %
$(ii)$ formulate or adapt a mathematical conjecture resembling the
informal question, $(iii)$ (dis)prove the conjecture.
Some questions are about approximating, implementing, and testing
various ideas and concepts.
Technically, many questions are on (the {\em interface} between) and
exploit {\em techniques} used in {\em (algorithmic) information
theory}, {\em machine learning}, {\em Bayesian statistics}, {\em
(adaptive) control theory}, and {\em reinforcement learning}.

\paradot{Feasibility, difficulty, and interestingness of the open questions}
I concentrated on questions whose answers probably help to develop
the foundations of universal induction and UAI. Some problems are
very hard, and their satisfactory solution worth a Nobel prize or
Turing award, e.g.\ problem \CGT. I included those questions that
looked promising and interesting at the time of writing this
article.
In the following I try to estimate their relative feasibility,
difficulty, and interestingness:

\begin{itemize}
\item Problems roughly sorted from most important or interesting to least: \\
\CGT,
\PNTM,\DPPM,
\OMUO,\DEE,
\IBR,\IOP,\PAIXI,\OIA,\CPCUH,\CRNCE,
\IZP,\IGRUE,\IRI,\IOEUP,\PSB,\OGBM,\SAEA,\SARV,\CRL,\CPOST,
\PGMCC,\OLEC,\SAIXI,\DGSPM,
\PMLC,\PLCBDM,\SPD.
\item Problems roughly sorted from most to least time consuming: \\
\IBR,\PNTM,\OIA,\CGT,\DEE,
\IGRUE,\OLEC,\OGBM,\SAIXI,\SAEA,\CRL,\CRNCE,\DPPM,
\PMLC,\PGMCC,\PAIXI,\OMUO,\SPD,\SARV,\CPCUH,\CPOST,
\IRI,\IOP,\PSB,\PLCBDM,\DGSPM,
\IZP,\IOEUP.
\item Problems roughly sorted from hard to easy: \\ \indent %
\PNTM,
\SAIXI,\CGT,
\IBR,\PMLC,\PAIXI,\OLEC,\SAEA,\DPPM,
\IGRUE,\PSB,\OMUO,\OGBM,\SARV,\CPCUH,\CRNCE,
\PGMCC,\CPOST,\DGSPM,
\IRI,\IOP,\PLCBDM,\OIA,\SPD,\DEE,
\IZP,\IOEUP,\CRL.
\end{itemize}
These rankings hopefully do not mislead but give the interested
reader some guidance where (not) to start.
The final paragraphs of this article are devoted to the role UAI
plays in the grand goal of AI.

\paradot{Other approaches to AI}
There are many fields that try to understand the phenomenon of
intelligence and whose insights help in creating intelligent
systems: {\em Cognitive psychology} and {\em behaviorism} \cite{Solso:07},
{\em philosophy} of mind \cite{Chalmers:02,Searle:05}, {\em neuroscience}
\cite{Hawkins:04}, {\em linguistics} \cite{Hausser:01,Chomsky:06},
{\em anthropology} \cite{Park:07}, {\em machine learning}
\cite{Sutton:98,Bishop:06}, {\em logic} \cite{Turner:84,Lloyd:87},
{\em computer science} \cite{Tettamanzi:01,Russell:03}, {\em biological
evolution} \cite{Kardong:07}, and others.
In computer science, most AI research is {\em bottom-up}; extending and
improving existing or developing new {\em algorithms} and increasing
their range of applicability; an interplay between experimentation
on toy problems and theory, with occasional real-world applications.
The agent perspective of AI \cite{Russell:03} brings some order and
unification in the large variety of problems the fields wants to
address, but it is only a framework rather than a complete theory.
In the absence of a perfect (stochastic) model of the environment,
machine learning techniques are needed and employed.
Apart from AIXI, there is no general theory for learning agents.
This resulted in an ever increasing number of {\em limited models and
algorithms} in the past.

\paradot{The information-theoretic approach to AI}
Solomonoff induction and AIXI are mathematical {\em top-down} approaches.
The price for this generality is that the full models are
computationally intractable, and investigations have to be mostly
theoretical at this stage.
From a different perspective, UAI strictly {\em separates
the conceptual and algorithmic AI questions}.
%
Two analogies may help: Von Neumann's optimal minimax strategy
\cite{VonNeumann:44} is a conceptual solution of zero-sum games, but
is infeasible for most interesting zero-sum games. Nevertheless most
algorithms are based on approximations of this ideal.
In physics, the quest for a ``theory of everything'' (TOE) lead to
extremely successful unified theories, despite their computational
intractability \cite{Green:87,Greene:00}.
The role of UAI in AI should be understood as analogous to the role
of minimax in zero-sum games or of the TOE in physics.

\paradot{Epilogue}
As we have seen, algorithmic information theory offers answers to
the following two {\em key scientific questions:} (1) The
problem of {\em induction}, which is what science itself is mostly
about: Induction $\approx$ finding regularities in data $\approx$
understanding the world $\approx$ science. (2) Understanding {\em
intelligence}, the key property that distinguishes humans from
animals and inanimate things.

This modern {\em mathematical} approach to both questions (1) and
(2) is quite different to the more traditional philosophical,
logic-based, engineering, psychological, or neurological approaches.
Among the few other mathematical approaches, none captures rational
intelligence as completely as the AIXI model does.
Still, a lot of questions remain open. Raising and discussing them
was the primary focus of this article.

Imagine a {\it complete} {\em practical} solution of the AI problem
(by the next generation or so), i.e.\ systems that surpass human
intelligence. This would {\em transform society} more than the
industrial revolution two centuries ago, the computer last century,
and the internet this century. Although individually, some questions
I raised seem quite technical and narrow, they derive their
significance from their role in a truly outstanding scientific
endeavor. As with most innovations, the social benefit of course
depends on its benevolent use.

\addcontentsline{toc}{section}{\refname}
\begin{small}\def\baselinestretch{0.93}
\newcommand{\etalchar}[1]{$^{#1}$}

\end{small}

\end{document}